\documentclass{article}
\usepackage{arxiv}
\usepackage[numbers]{natbib}
\usepackage[utf8]{inputenc} 
\usepackage[T1]{fontenc}    
\usepackage{hyperref}       
\usepackage{url}            
\usepackage{amsfonts}       
\usepackage{nicefrac}       
\usepackage{microtype}      
\usepackage{graphicx}
\usepackage{doi}
\usepackage{amssymb}

\usepackage{microtype}
\usepackage{booktabs} 
\usepackage{subfig}
\usepackage[utf8]{inputenc}
\usepackage{lineno}
\usepackage{multirow}
\modulolinenumbers[10]

\usepackage{amsmath}
\usepackage{amsxtra}
\usepackage{amstext}
\usepackage{amsfonts}
\usepackage{mathtools}
\usepackage{amsthm}
\usepackage{acronym}

\usepackage[capitalize,noabbrev]{cleveref}

\usepackage[textsize=tiny]{todonotes}

\usepackage{tikz}
\usetikzlibrary{shapes.multipart, backgrounds,calc}

\usepackage{lipsum}
\usepackage{blindtext}
\usepackage{todonotes}
\usepackage{placeins}

\usepackage{enumitem}

\newcommand{\EE}[2][]{\mathbb{E}_{#1}\left[#2\right]} 
\newcommand{\IR}{\mathbb{R}} 

\newcommand{\ie}{\emph{i.e.}}

\DeclareMathOperator*{\argmax}{\mathrm{argmax}}

\makeatletter
\newtheoremstyle{indented}
  {1em}
  {1em}
  {\addtolength{\@totalleftmargin}{2em}
   \addtolength{\linewidth}{-4em}
   \parshape 1 2em \linewidth \itshape}
  {}
  {\bfseries}
  {}
  {.5em}
  {}
\makeatother
\theoremstyle{indented}

\newtheorem*{question*}{Research Question}

\acrodef{HRI}{Human-Robot Interaction}
\acrodef{MDP}{Markov decision problem}
\acrodef{POMDP}{partially observable Markov decision problem}
\acrodef{PoLMDP}{policy legible Markov decision problem}
\acrodef{L-MDP}{legible Markov decision problem}
\acrodef{I-POMDP}{interactive POMDP}
\acrodef{KL}{Kullback-Leibler}
\acrodef{IRL}{inverse reinforcement learning}
\acrodef{RL}{reinforcement learning}
\acrodef{AIA}{artificial intelligent agent}
\acrodef{AI}{artificial intelligence}
\acrodef{IVA}{intelligent virtual agent}

\title{``Teammates, Am I Clear?'': Analysing Legible Behaviours in Teams}

\author{ \href{https://orcid.org/0000-0002-0470-0739}{\includegraphics[scale=0.06]{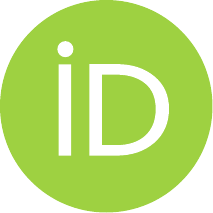}\hspace{1mm}Miguel Faria}\\
	SARDINE, Instituto de Telecomunicações \& Instituto Superior Técnico\\
	Lisbon, Portugal\\
	\texttt{miguel.faria@tecnico.ulisboa.pt}\\
	\And
	Francisco S. Melo \\
	INESC-ID \& Instituto Superior Técnico\\
	Lisbon, Portugal\\
	\texttt{fmelo@inesc-id.pt} \\
	\And
	Ana Paiva\\
	INESC-ID \& Instituto Superior Técnico\\
	Lisbon, Portugal\\
	\texttt{ana.paiva@inesc-id.pt} \\
}

\begin{document}

\maketitle 

\begin{abstract}
In this paper we investigate the notion of \emph{legibility} in sequential decision-making in the context of teams and teamwork. There have been works that extend the notion of legibility to sequential decision making, for deterministic and for stochastic scenarios. However, these works focus on one agent interacting with one human, foregoing the benefits of having legible decision making in teams of agents or in team configurations with humans. In this work we propose an extension of legible decision-making to multi-agent settings that improves the performance of agents working in collaboration. We showcase the performance of legible decision making in team scenarios using our proposed extension in multi-agent benchmark scenarios. We show that a team with a legible agent is able to outperform a team composed solely of agents with standard optimal behaviour.
\end{abstract}



\section{Introduction}
\label{sec:intro}

\Acp{AIA} are becoming more present in society, varying from \acp{IVA}~\cite{calvo2022iva} to embodied physical agents~\cite{melo2019aim}. The presence of \acp{AIA} in society has made them more than mere tools; they started to be seen as peers and partners of humans interacting with them. Humans have started to rely on \acp{AIA} for help in healthcare situations~\cite{alves15icsr}, in education~\cite{chandra2016roman}, entertainment~\cite{correia2017social}, among others. All of these applications require correct communication for humans to understand the agent's goals and intentions. Communication can be done explicitly, for example through speech, or implicitly for example, through body motions~\cite{bauer2008ijhr}. Implicit communication is extremely important to create natural interactions, because humans naturally interpret behaviours, actions and movements from each other to have a better understanding of the underlying intentions~\cite{gildert2018frontiers}. 

Imagine the task of collaborative assembly of a cabinet with drawers. To show that we are mounting the cabinet's frame first, we can first collect all the pieces required to mount the frame before starting to assemble it, leaving it for our partner to mount the drawers. This shows how sequencing the actions by first collecting all the pieces required for the task and only then assemble the frame, conveys more information about our objective than collecting the pieces as needed while mounting the frame.



Motivated by situations like the previous example, in this work we explore the following question:
\begin{center}
	\emph{``Does the use of legible decisions improve the efficiency of a team in collaborative tasks?''}
\end{center}

The notion of \textit{legibility} was introduced by \citet{dragan2013hri} and focused on creating expressive movements. Legibility measures how much a user is able to discern the goal of a robot from an observed snippet of movement. A legible movement is characterized not by its {\em efficiency} in reaching the goal, but by its {\em distinctiveness}, \ie, how much it is able to disambiguate the actual goal of the movement from other potential goals. 
%
%
Legibility has been widely explored in human-robot interaction to improve a robot's expressiveness through movement, both in single-user~\cite{dragan2015hri} and multi-user~\cite{faria21roman} interactions. 

Recently, several works have explored the application of legibility outside of robotic movement, motivated by improving the transparency and explainability of autonomous systems~\cite{anjomshoae19aamas}. The works of \citet{habibian22ral}, \citet{macnally2018aamas}, \citet{kulkarni2019aaai} and \citet{masters2019aamas} explored the effects of legible behaviours in scenarios where the outcomes of an agent's actions are deterministic, much as in the planning setting of \citet{dragan2013hri}. \citet{habibian22ral} focus on legibly allocating subtasks to teams composed of humans and robots, creating fair and easier to understand divisions of labour; \citet{macnally2018aamas}, \citet{kulkarni2019aaai} and \citet{masters2019aamas} focus on the decision making of an agent in deterministic scenarios, to improve the agent's expressiveness and create plans that are easier to understand. On the other hand, the works of \citet{miura2021roman} and \citet{faria22arxiv} focus on legible behaviours when planning under uncertainty, where the outcome of an agent's action is stochastic and thus planning must account for that stochasticity. Both works present two frameworks to generate sequences of legible actions for single-agent scenarios. \citet{miura2021roman} focus on an approach similar to \acp{POMDP}, while \citet{faria22arxiv} present an approach similar to \acp{MDP}. More recently, \citet{liu2025ai} proposed LI-POMDP, an approach for legible decision making in multiagent environments. LI-POMDP, like the work of \citet{miura2021roman}, uses a \ac{POMDP}-like formalism and assumes all agents act legibly to directly derive a legible decision policy from interaction with the environment, without first deriving an optimal policy.

Despite their focus on developing agents capable of these legible behaviours, most works on legible decision making focus on single-agent scenarios and on how one agent can be more expressive about its intentions to a single human user. There is not much work on the impact of legible actions in collaborative efforts, where previous works of \citet{mavrogiannis18hri} and \citet{faria21roman} have shown that legibility has a great impact. We address this gap by exploring the impact of using legible behaviours in multi-agent scenarios, namely in scenarios that require collaboration between teammates. So, we evaluate the team performance and intention communication between teammates in teams where agents use only optimal behaviours or a combination of legible and optimal behaviours. Our experiences use two common scenarios in the \ac{RL} literature: the LB-Foraging benchmark scenario~\cite{christianos2020shared} and the Pursuit-Evasion problem. Our experiments show that teams where agents are capable of legible behaviours improve team efficiency in collaborative tasks.

\section{Background}
\label{sec:background}

This section introduces key concepts and notation used in the remainder of our work.

\subsection{Markov decision problems}
\label{subsec:mdps}

A {\em Markov decision problem} (MDP) is a model for sequential decision making in stochastic environments. A \ac{MDP} $M$ is defined as a tuple $\left<X, A, P, r, \gamma \right>$, with $X$ the state space; $A$ the action space; $P$ the state transition probabilities, where $P(y\mid x,a)$ indicates the probability of moving from state $x$ to state $y$ upon executing action $a$; $r:X\times A\to\IR$ is the reward function; $\gamma\in[0,1)$ is a discount factor, indicating the relative importance of future rewards against present rewards. An \ac{MDP} describes the sequential interaction between an agent and its environment where, at each time step $t$, the agent observes the state of the environment, $X_t\in X$, performs an action $A_t\in A$ and, as a result, observes a reward $R_t\in\IR$, where
\begin{equation*}
\EE{R_t\mid X_t=x,A_t=a}=r(x,a).
\end{equation*}
Solving a \ac{MDP} amounts to computing an {\em optimal policy}, $\pi^*$. A policy, $\pi$, is a mapping from states to actions describing the action for the agent to take in each state, and we can define the {\em value} associated with a policy as
\begin{equation*}
	v^\pi(x)=\EE{\sum_{t=0}^\infty\gamma^tR_t\mid X_t=x},
\end{equation*}
where $X_t$ and $R_t$ are the state and reward at time step $t$, respectively. The optimal policy is such that $v^{\pi^*}(x)\geq v^\pi(x)$ for all $x\in X$ and all policies $\pi$. The value associated with the optimal policy is denoted $v^*$, and we define the {\em optimal $Q$-function} as
\begin{equation*}
	q^*(x,a)=r(x,a)+\gamma\sum_{y\in X}P(y\mid x,a)v^*(y).
\end{equation*}
The optimal $Q$-function can be computed in polynomial time using dynamic programming~\cite{papadimitriou87mor}, and the optimal policy can be computed from $q^*$ simply as $\pi^*(x)=\argmax_{a\in A}q^*(x,a)$, for all $x \in X$.

%
%
%
%
%

\subsection{Policy legible \acp{MDP}}
\label{subsec:legibility}

Legibility describes how readable an action or movement's objective is, and is inspired by the principle of rational action~\cite{popper1976myth}, which states that a ``{\em rational agent will act efficiently and justifiably to achieve its goals.}'' The legibility of an action is quantified as the probability of a human assigning one specific objective, $g$, to the agent's intentions, after observing a snippet of the agent's trajectory, $\xi_{x_0 \to x_t}$, with $x_0$ the agent's starting state and $x_t$ the robot's state at time $t$. The goal inference can be defined as 
\begin{equation}
	\label{eq:legible-inference}
	\mathcal{I}_L (\xi_{x_0 \to x_t}) = \argmax_{g\in G} P(g | \xi_{x_0 \to x_t}),    
\end{equation}
where $G$ is the set of possible goals of an agent. Using Bayes' Rule
\begin{equation}
	\label{eq:legibility-target-prob}
	P(g | \xi_{x_0 \to x_t}) \propto P(\xi_{x_0 \to x_t} | g) P(g),
\end{equation}
with $P(g)$ the prior on the goals and $P(\xi_{x_0 \to x_t} | g)$ models the probability of the observed trajectory snippet being observed when the agent progresses towards goal $G$. In the original work of \citet{dragan2013hri}, the latter is expressed as a maximum entropy distribution in the form
\begin{equation*}
	P(\xi_{x_0 \to x_t} | g)=\frac{\exp(-C(\xi_{x_0 \to x_t})-C(\xi^*_{x_t \to g}))}{\exp(-C(\xi_{x_0\to g}))},
\end{equation*}
where $\xi^*_{x_t \to g}$ denotes the optimal trajectory from the robot's pose at time $t$ to the goal pose $g$ and $C(\xi)$ is the cost associated with trajectory $\xi$.

Expanding on the definition from \citet{dragan2013hri}, \citet{faria22arxiv} proposed the {\em Policy Legible \ac{MDP}} (\acs{PoLMDP}) framework that uses the formalism of \acp{MDP} to generate legible behaviours. A \acs{PoLMDP} is defined in the context of environments with $N$ different objectives, each with a different reward function $r_n$, $n = 1, 2, \dotsc, N$, and thus describing a different \ac{MDP} -- $\ac{MDP}_1, \ac{MDP}_2, \dotsc, \ac{MDP}_N$. In a \acs{PoLMDP}, the agent tries to maximize the legible reward $r_{leg}$, which measures an action's legibility. According to \citet{faria22arxiv}, $r_{leg_n}$ evaluates the likelihood of executing action $a$ in state $x$ to reach objective $n$, in opposition to executing the same action at the same state to reach another of the $N$ possible objectives. Thus, $r_{leg_n}$ is defined, for a reward function $r_n$, as follows
\begin{equation}\label{eq:action-legibility}
	r_{\rm leg}(x, a) = P(r_n\mid (x, a)),
\end{equation}
where
\begin{equation*}
	P(r_n\mid (x, a)) \propto P((x, a)\mid r_n) P(r_n).
\end{equation*}
With this representation, the probability of executing an action $a$ in state $x$ is not influenced by possible states and actions that preceded the current state. This way, an agent needs only to observe the current state-action pair to infer the robot's intentions. Considering a uniform distribution as the prior on the probability of observing each goal, the previous expression is can be simplified as
\begin{equation}
	\label{eq:legible-reward}
	r_{\rm leg}(x, a) = P((x, a)\mid r_n).
\end{equation}
The policy derived by optimizing the underlying expected legible reward defines an optimal legible policy, $\pi_{leg}^*$.


\section{Legible decision in teamwork scenarios}
\label{sec:legible-teams}

\begin{figure}[t]
	\centering
	\includegraphics[width=.65\linewidth]{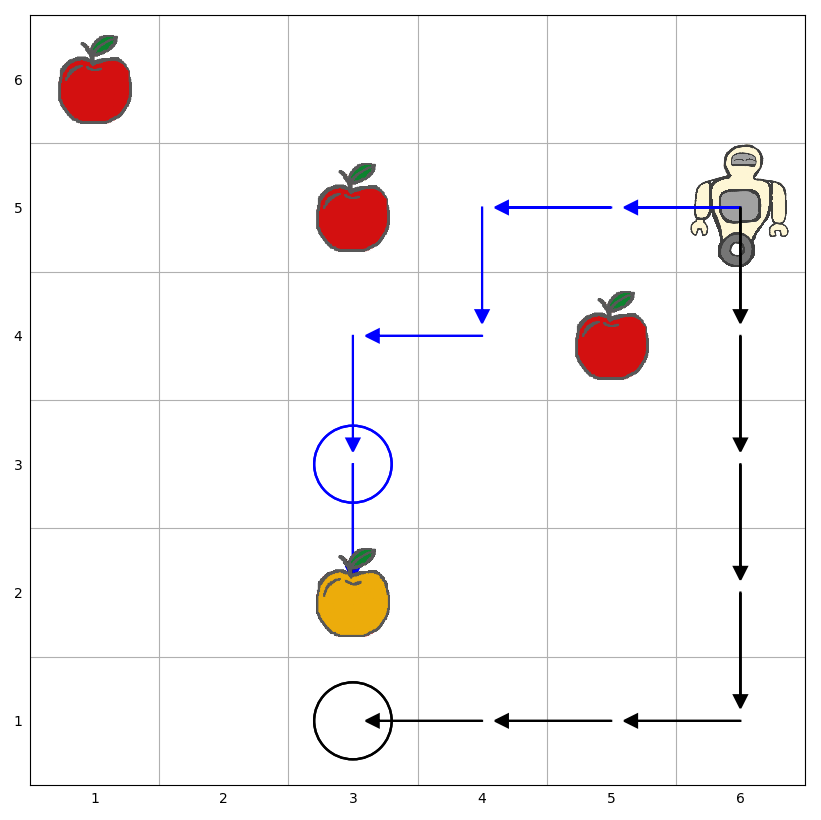}
	\caption{Example of the difference between optimal and legible decision-making while picking up food in a level-based apple foraging scenario. The agent's current target is the gold apple. The blue arrows indicate an optimal sequence of actions and the black arrows a legible sequence of actions. The circles represent the agent trying to pickup the apple.}
	\label{fig:example-movement}
\end{figure}

Solving an \ac{MDP} consists of computing the an optimal policy $\pi^*$. However, the optimal policy focuses on actions that yield better long-term rewards and does not guarantee that, in an interaction, an observer can easily understand the agent's intentions. Figure~\ref{fig:example-movement} shows two decision sequences for an agent to reach the gold apple: in blue is an optimal sequence of actions to pickup the gold apple; while in black a sequence of legible actions obtained using the \acs{PoLMDP} framework. As we can observe in Figure~\ref{fig:example-movement}, the legible sequence of actions, moves the agent as far away as possible from other apples in the environment, instead of following the most direct path that goes closer to more apples. Legible decision making, thus, aims at improving an agent's transparency besides improving its underlying reward; making the agent's intentions clearer to an outside observer that can only observe the outcome of the agent's actions.

In this work we use the framework of \emph{\acs{PoLMDP}}~\cite{faria22arxiv} for legible decision making, despite more recent approaches like \citet{liu2025ai}'s LI-POMDP that allows to derive directly a legible policy from interacting with the environment. We use \acs{PoLMDP} because it does not have the assumption that all agents are legible agents, like it happens with the LI-POMDP framework, and \acs{PoLMDP} has been shown to achieve solutions faster than other state-of-art approaches, while maintaining the same level of legibility~\cite{faria22arxiv}. 

\subsection{Multi-agent legible decision making}

In multi-agent scenarios, the dynamics are influenced by the combination of actions of all the agents. The action space, $A$, becomes a joint action space $A = A_1 \times A_2 \times \dotsc \times A_N$ with $N$ the number of agents. By consequence, the transition probabilities $P$ and reward function $r$ also have to account for the joint actions, instead of individual agent actions. This makes the decision process more complex than in single-agent scenarios, because the optimal policy $\pi^*$ needs to account for other agents executing varying actions. Thus, an agent's $\pi^*$ needs to model a best response to the other agents' actions.


A simple solution for multi-agent optimal behaviour in teamwork scenarios is to consider the agents' actions as centralized and, thus, the optimal policy $\pi^*$ models a centralized joint behaviour of the agents. Modelling the agents' actions as a centralized joint policy allows for better coordination between team members, as a single action prescribes the behaviour of all agents. However, this centralized behaviour, leads to the agents losing independence from each other, and assumes a central entity has access to all the agents' observations and control over all agent; which does not correctly model how a real interaction occurs. In the real world, each agent may act independent of the other agents, without a central decision process deciding each agent's actions, and agents do not share their observations with each other. So, we adapt the original formulation of a \acs{PoLMDP} policy, defined for single-agent scenarios, to multi-agent cooperative scenarios using a centralized training and decentralized execution approach, better resembling how agents in the real world would interact and perceive the environment.


In order to adapt \acs{PoLMDP} from single-agent to multi-agent scenarios, we use a value-based centralized training and decentralized execution approach. During training, each agent learns its own decision policy; having access to the other agents' actions, so it can evaluate the utility of a joint action for each given state. At execution time, each agent acts individually using the learned individual decision model. We use duelling double deep Q-networks~\cite{wang2016icml, hasselt2016aaai} as each agent's decision model and for evaluating the joint action's utility we use a value decomposition network approach~\cite{sunehag2017value}.

We learn the multi-agent \acs{PoLMDP} legible policy in a two stage approach. We start by learning the optimal joint decision policy for the environment's underlying reward function, $\pi^*$, and then use the learned optimal policy to train the optimal legible policy, $\pi^*_{leg}$. Considering an environment with $N$ controllable agents, of which $L$ are legible agents, we start by training $\pi^*$ considering all $N$ agents optimize the environment's reward function. With the joint $\pi^*$ learned, we fix $N - L$ agents to follow $\pi^*$, and the remaining $L$ agents learn $\pi^*_{leg}$. To learn $\pi^*_{leg}$, the $L$ agents optimize the legible reward, described in the Background section, while interacting with the agents following $\pi^*$. Thus, the learned legible decision policy is a best response to the actions of the other optimal teammates. This training process allows to flexibly adapt the training regimen of new legible policies when the ratio of legible to optimal agents in the environment changes. Also, we can train legible agents that maintain independent agency but are not indifferent to the other agents in the environment, thus more closely reflecting how intelligent agents behave in real world scenarios.

%
%



\section{Legible Teamwork Experiments}
\label{sec:study}

This work focus on the question:
\begin{center}
	\emph{``Does the use of legible decisions improve the efficiency of a team in collaborative tasks?''}
\end{center}

To support the exploration of our problem we postulate the following working hypotheses:
\begin{enumerate}[label={\bf H\arabic*}]
	\item \emph{Team performance will increase when agents use legible behaviours}.
	\item \emph{Agents observing legible behaviours will be able to understand other agents' intentions faster}.
\end{enumerate}

\begin{figure}[t]
    \centering
    \subfloat[LB foraging scenario]{
        \includegraphics[width=.4\linewidth]{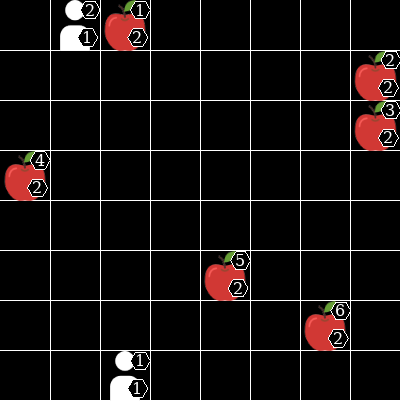}
        \label{fig:lb-example}
    } \qquad
\subfloat[Pursuit-evasion scenario]{
        \includegraphics[width=.4\linewidth]{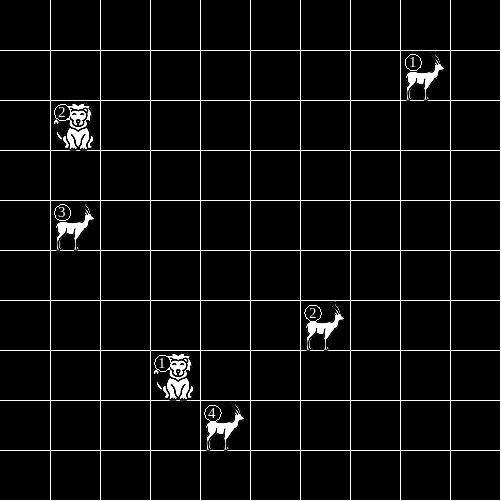}
        \label{fig:pursuit-example}
    }
    \caption{LB foraging and pursuit-evasion examples. In~\ref{fig:lb-example} we have an example of lb-foraging with two agents and six food items spread in the environment. All food items are level two foods and both agents are level one agents. In~\ref{fig:pursuit-example} we have an example of pursuit-evasion with two hunters and four preys spread in the environment.}
    \label{fig:scenario-examples}
\end{figure}

In this section, we present two experiments where teams of agents have to fully cooperate to finish the task. In the first experiment we used the level-based foraging scenario (LB-Foraging) from \citet{papoudakis2021benchmarking}; in the second, we used the pursuit-evasion scenario. In the LB-Foraging scenario, exemplified in Figure~\ref{fig:lb-example}, a team of agents has to pick all food items in the environment, where each agent and food item has a level associated with it. In order to pick a food item, one or more agents have to collaborate so that the sum of their levels is greater than or equal to the level of the food item they are trying to pick. In the pursuit scenario, exemplified in Figure~\ref{fig:pursuit-example}, there are two teams -- hunters and preys -- and the objective is for the team of hunters to capture all the preys before the time runs out. For a prey to be considered captured, it must be surrounded by a predefined number of hunters. 

Our study featured two conditions, reflected in the team compositions. Each team was composed by one leader agent and one follower agent. In the first condition, which henceforth we will refer as the \textit{optimal condition}, the leader agent followed an optimal policy as the decision policy. In the second condition, which henceforth we will refer as the \textit{legible condition}, the leader followed a \acs{PoLMDP} policy. In both conditions, the follower agent used an optimal policy as the decision policy.

%

\subsection{Team Composition}
\label{subsec:team-comp}

Each team was composed by a leader and a follower agent: the leader knew the sequence of objectives to fulfil, and was responsible to implicitly communicate it through its actions; and the follower agent had to infer the sequence of objectives from the leader's actions. Both agents used a \ac{MDP} formalism in their decision making process; however, the implementations differed depending on the agent's role. The follower agent, having to infer the sequence of objectives, used a Bayesian inference process, as in GIRL~\cite{lopes09ecml}, to infer the next objective in the sequence; and, for decision making, used a standard \ac{MDP} that optimized the reward function of the environment. The leader agent, being responsible for transmitting information, did not have to infer the next objective and used one of two types of \acp{MDP} for decision making. In the optimal condition, the leader used a standard \ac{MDP} that optimized the reward function of the environment; while in the legible condition, the leader used a \acs{PoLMDP} that optimized the corresponding legible reward. 

\subsection{Level-based foraging experiments}
\label{subsec:lb-foraging}

The LB-Foraging scenario focused on exploring the impact of a legible team leader in pure collaborative scenarios. This environment resembles real-world tasks where a team of agents has to interact with objectives that do not change state without outside interference. To increase the focus on the communication of intentions, we forced the agents to all have the same level and the food items' levels to be equal to the sum of the agents' levels. So, no agent could pick a food item alone and they had to come together to pick each food item. This configuration of the environment avoided agents to learn strategies where just a small set of agents could solve most of the task alone, which would remove the focus from correct transmission of intentions.


These experiments used a configuration with two agents and six food items spread in the environment. We varied the field size to test team's performance with an increasing number of state spaces. This allowed to understand the impact of legibility when the agents have more freedom of movement and the decision processes are longer to reach the goal. We set up 7 different grid sizes -- $5\times5$, $6\times6$, $7\times7$, $8\times8$, $10\times10$, $15\times15$, and $20\times20$ squares -- and for each grid size we chose 8 possible food item locations, of which six were uniformly sampled in each run. We decided to have 8 possible food locations for each grid size to force configurations to have a mix of isolated and clustered items. With such a mix, we always had items that were easier to identify as the objective and others that were harder. By sampling 6 out of the 8 possible locations, we also added diversity to the scenarios the agent teams faced, leading to more thorough testing.

\subsubsection{Metrics}
\label{subsubsec:metrics-lb}

We measured the impact of legibility in team performance by the number of steps needed for each team to finish the task. The impact of legibility in intention transmission was measured by the average number of steps the followers needed to correctly infer the leader's next objective. We ran 250 simulations for each combination of condition and grid size. To guarantee that both conditions were exposed to the same testing conditions, and one condition did not benefit from easier capture sequences, we pre-sampled the initial environment configuration and the corresponding food capture sequence, for all 250 simulations. Thus, for each grid size, both conditions had the same testing configurations. Finally, we added a time limit of 600 steps to complete the objective sequence. 

\subsubsection{Evaluation Results}
\label{subsubsec:results-lb}

After running all simulations we got 500 samples for each grid size. We then validated the results obtained and ended up removing one sample from both the $ 5 \times 5 $ and $ 6 \times 6 $ grids and 5 from the $ 7 \times 7 $ grid. We removed these samples because all of them reported failures: when analysing the trial's logs, we observed that the failures were due to an environment error, where the pick action was not correctly recognized preventing the task from progressing. Thus, we removed the corresponding samples for both conditions, resulting in 249 samples for each condition in both the $ 5 \times 5 $ and $ 6 \times 6 $ grids and 245 samples for each condition in the $ 7 \times 7 $ grid.

After validating the results, we conducted a normality test on the results, which showed that their distribution significantly deviated from the normal distribution. With this in mind, we only conducted non-parametric tests, namely we used the Mann-Whitney U Test.

\begin{figure}[!t]
	\centering
	\includegraphics[width=.75\linewidth]{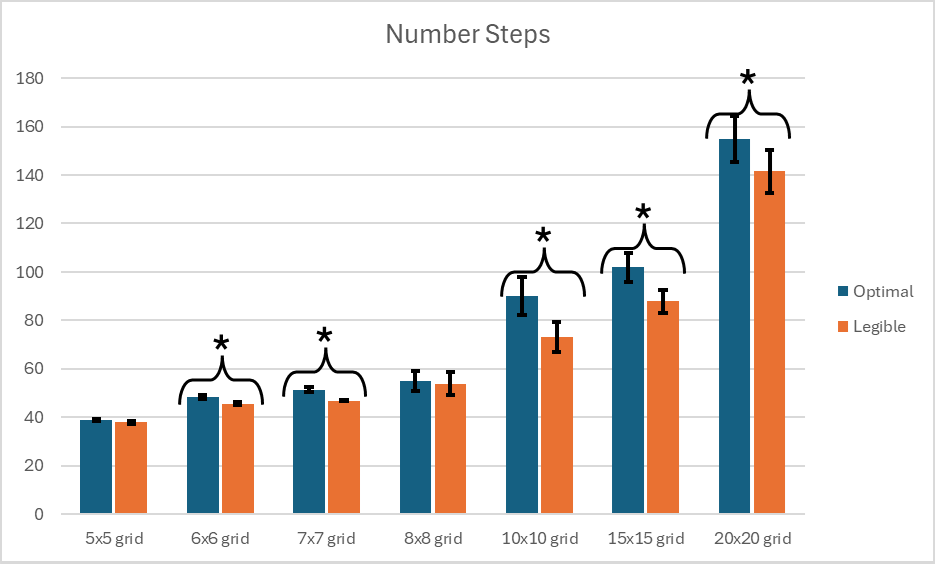}
	\caption{Results for the average number of steps to pick all the food items in the world, in black bars are the standard error bars. (*$p < 0.05$)} 
	\label{fig:results-goal-lb}
\end{figure}

Figure~\ref{fig:results-goal-lb} shows the average number of steps needed for each team to pick all food items, for the different grid sizes. The results show that the agents in the legible condition achieved a better or equal performance than the agents in the optimal condition. 
Only on the $ 5 \times 5 $ and $ 8 \times 8 $ grids the performance of both teams was similar, with the optimal condition taking on average $39$ and $55$ steps, respectively, to pick all the food items and the legible condition taking on average $38$ and $54$ steps, respectively. For the other grid sizes, the legible teams needed significantly less steps than the optimal teams to collect all food items. These conclusions were validated by a Mann-Whitney U test that showed significant differences for the grid sizes: for the $ 6 \times 6 $ grids $U = 26204, p = 0.03$; for the $ 7 \times 7 $ grids $U = 24110.5, p < 0.001$; for the $ 10 \times 10 $ grids $U = 25820.5, p < 0.001$; for the $ 15 \times 15 $ grids $U = 25893, p < 0.001$; and for the $ 20 \times 20 $ grids $U = 26799, p = 0.006$.

\begin{figure}[!t]
	\centering
	\includegraphics[width=.75\linewidth]{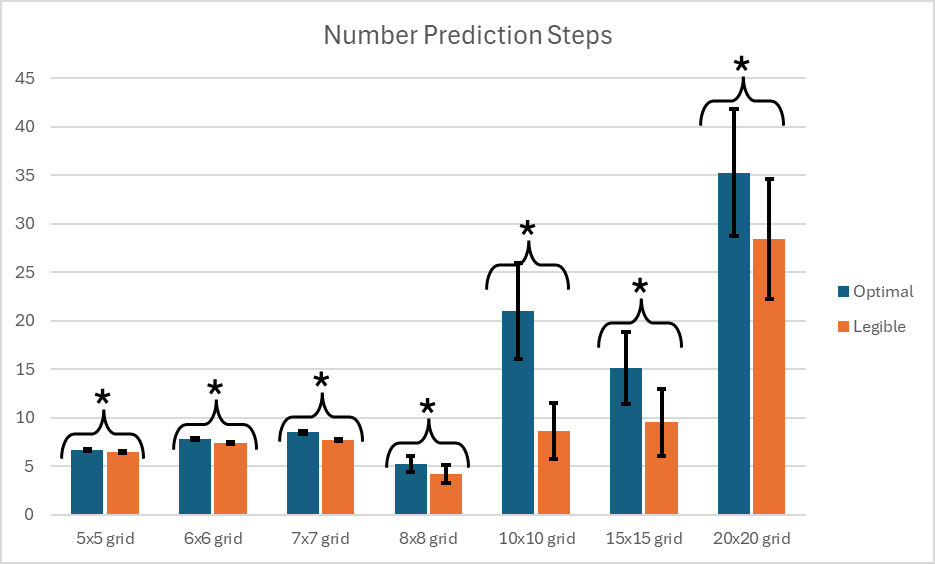}
	\caption{Results for the average number of steps to correctly infer the next food item to be picked, in black bars are the standard error bars. (*$p < 0.05$)} 
	\label{fig:results-inference-lb}
\end{figure}


Figure~\ref{fig:results-inference-lb} shows the average number of steps needed for the follower agent to correctly infer the next food item to be picked, for the different grid sizes. The results show that for all grid sizes legible behaviours led the follower to infer the next food item faster than in the optimal condition. In the case of grid sizes up to $ 8 \times 8 $, the legible condition got an average improvement between half and a full step per food item. For the larger grids, the differences were starker, varying between $\simeq8$ and $\simeq17$ steps per food item. To understand if the differences between the conditions had statistical significance, we conducted a Mann-Whitney U test, which showed a significant difference for all grid sizes: for $ 5 \times 5 $ grids $U = 26755.5, p = 0.08$; for $ 6 \times 6 $ grids $U = 24294, p < 0.001$; for $ 7 \times 7 $ grids $U = 23058, p < 0.001$; for $ 8 \times 8 $ grids $U = 17693, p < 0.001$; for $ 10 \times 10 $ grids $U = 15195.5, p < 0.001$; for $ 15 \times 15 $ grids $U = 13129.5, p < 0.001$; and for $ 20 \times 20 $ grids $U = 15732.5, p < 0.001$.


\subsubsection{Discussion}
\label{subsubsec:discussion-lb}

The results in Section~\ref{subsubsec:results-lb} allow us to conclude that the performance teams with a legible leader was higher than teams with an optimal leader. On average, teams with a legible leader needed significantly less steps to pick all food items than teams only with optimal agents; which is aligned with our hypothesis \textbf{H1} that postulated that using legible behaviours would increase team performance.


The second conclusion from the results is that legible behaviours allow a follower agent to infer the current objective and, consequently, the objective sequence faster. This result is in line with our hypothesis \textbf{H2} that postulated that agents observing legible behaviours would understand the intentions of other agents faster. Besides validating \textbf{H2}, this results also supports hypothesis \textbf{H1}. This is due to agents requiring less steps to understand the current objective, leading to them locking into follow a single decision policy faster and wasting less time following wrong policies.



Finally, a third conclusion emerges from analysing the evolution of differences between the two conditions in both task completion steps and steps required to infer the next objective. The results show that these differences tend to widen as the state space increases. This suggests that in larger state spaces, a legible agent can exploit the additional flexibility to identify trajectories that enhance team efficiency. This is supported by Figure~\ref{fig:results-inference-lb}, which shows that in larger grids, the difference in steps needed to infer the next food item increases from $\simeq1$ step to as many as seventeen. These results indicate that a legible agent can leverage the greater range of options in less constrained environments to improve team performance by providing more informative behaviour to teammates.

\subsection{Pursuit-evasion experiments}
\label{subsec:pursuit-evasion}

The pursuit-evasion experiments focused on exploring the impact of legible decisions when multiple teams have different objectives. Thus, we can understand how legible decision making influences team efficiency in real world scenarios like search and rescue scenarios or team competitions, where different teams might have different -- even opposing -- objectives, such as in our case the hunters aim to capture the preys and the preys aim at evading.

For these experiments, we used a configuration with 2 hunters and 7 preys. We evaluated the hunter team performance in an increasing number of states by having 3 different grid sizes -- $ 10 \times 10 $, $ 15 \times 15 $, and $ 20 \times 20 $. The preys moved greedily, maintaining the biggest distance possible from all hunters in the environment. Using greedy prey behaviour has two main advantages: first, it closely resembles the behaviour of entities being pursued in real-world scenarios that try to escape the pursuer as fast as possible, without must long-term planing or coordination with others; second, as a comparison baseline, it is the simplest rational type of behaviour to implement and so is good to understand the impact of legible behaviours, without other side effects impacting the communication of intentions.

\subsubsection{Metrics}
\label{subsubsec:metrics-pe}

For the Pursuit-Evasion scenario we recorded both the number of steps needed to capture all preys and for the follower to correctly infer the next prey, without posterior wrong predictions. We ran 200 simulations for each combination of condition and grid size. To guarantee that both conditions were exposed to the same configurations, we pre-sampled the initial positions of both hunters and preys and the prey capture sequence. Finally, we gave a time limit of 800 steps for the hunter team to capture all the preys.

\subsubsection{Evaluation Results}
\label{subsubsec:results-pe}

With the simulation results, we conducted Kolmogorov-Smirnov normality tests that showed the results significantly deviated from a normal distribution. With this in mind, we only conducted non-parametric tests, namely we used the Mann-Whitney U test.

\begin{figure}[!t]
	\centering
	\includegraphics[width=.75\linewidth]{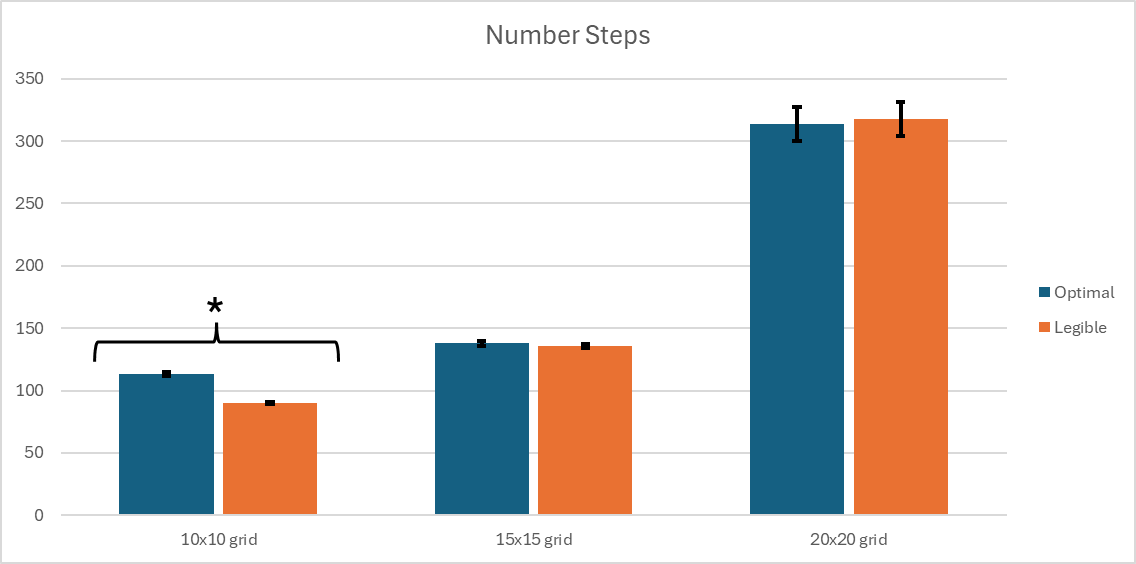}
	\caption{Results for the average number of steps needed to capture all the 7 preys, in black bars are the standard error bars. (*$p < 0.05$)} 
	\label{fig:results-goal-pursuit}
\end{figure}

Figure~\ref{fig:results-goal-pursuit} shows the average number of steps needed to capture all the preys, for each team configuration and grid size. The results show the legible team's equiparable or better performance to the optimal team. In the smaller grid size the legible team was noticeably better than the optimal team, with the legible team taking, on average, $\simeq90$ steps to capture all preys and the optimal team $\simeq113$ steps. For the two larger grids -- $ 15 \times 15 $ and $ 20 \times 20 $ -- both teams perform similarly, requiring, on average, $\simeq135$ steps to capture all preys in the $ 15 \times 15 $ grids and $\simeq315$ steps in the $ 20 \times 20 $ grids. These conclusions were validated using the Mann-Whitney U test that only showed significant differences in the $ 10 \times 10 $ grids $U = 8283.5, p < 0.001$.

\begin{figure}[!t]
	\centering
	\includegraphics[width=.75\linewidth]{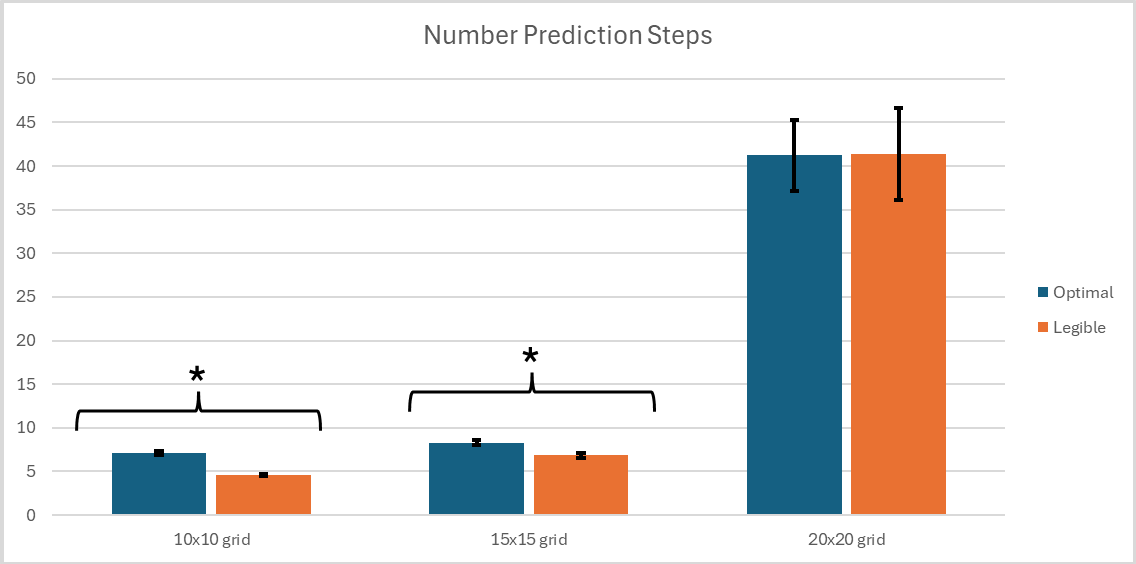}
	\caption{Results for the average number of steps needed for the follower to correctly identify the next prey to capture, in black bars are the standard error bars. (*$p < 0.05$)} 
	\label{fig:results-inference-pursuit}
\end{figure}

Figure~\ref{fig:results-inference-pursuit} shows the average number of steps needed for the follower agent to correctly infer the next prey to capture, for each team configuration and grid size. The results show that in the $ 10 \times 10 $ and $ 15 \times 15 $ grids legible behaviours led the follower agents to infer the next prey faster than optimal behaviours; with the followers in the legible condition showing, on average, an improvement between $\simeq1$ and $\simeq2.5$ inference steps per prey. For the $ 20 \times 20 $ grids both conditions led the follower to require a similar number of steps to infer the next prey: $\simeq41$ steps on average. To understand the significance of these differences, we conducted Mann-Whitney U tests for all grid sizes. For the $ 10 \times 10 $ and $ 15 \times 15 $ grids the tests reported a significance difference between the two conditions: for $ 10 \times 10 $ grids $U = 9083.0, p < 0.001$ and for $ 15 \times 15 $ grids $U = 15379.0, p < 0.001$.

\subsubsection{Discussion}
\label{subsubsec:discussion-pe}

The results in Section~\ref{subsubsec:results-pe} show that overall, in a task where multiple teams have different objectives, a team with a legible leader offers benefits over teams with optimal leaders. Focusing on the the number of steps to capture all the preys, these partially support hypothesis \textbf{H1} that postulated that using legible behaviours would increase team performance. Although in the smaller grid sizes legible behaviours lead to an increase in team performance, in the bigger grids the improvement was not verified. This contrasts with the results in Section~\ref{subsubsec:results-lb}, for the LB-Foraging scenario, , where the difference between the legible and optimal conditions was bigger in the bigger grids than in the smaller grids. 

Regarding the results for the steps needed to correctly infer the next prey, these are mostly aligned with hypothesis \textbf{H2} that postulated that observing legible behaviours would lead to agents infer objectives faster. From the results we can observe that only in the biggest grids hypothesis \textbf{H2} was not supported, which again contrasts with the results for the LB-Foraging experiments, where the most stark differences observed were in the bigger grids.

Overall, legible behaviours provided greater benefits in smaller grids, with diminished advantage in larger ones. One reason for this reverse trend might stem from the inherently competitive nature of pursuit-evasion, where hunters must act in a way that prevents the preys from escaping or clustering. In larger grids, optimal leaders’ direct actions toward capture leave less opportunity for prey to cluster, matching the performance of legible leaders, whose behaviour prioritizes signalling intent over capture efficiency.


\section{Final Remarks}
\label{sec:conclusions}

In this work we explored the problem of using legible behaviours in collaborative tasks, namely the utility of legible behaviours in tasks that require teamwork. Using \acs{PoLMDP}, a framework for legible decision making, we studied how a team composed by legible and optimal agents performed, in terms of time to solve the tasks and communicate intentions, when compared to a team composed only by optimal agents. 

We conducted a study to compare the performance of a team that used legible behaviours and one that did not, in the scenarios of LB-Foraging and Pursuit-Evasion. With the LB-Foraging scenario, we focused on exploring the utility of legible behaviours in pure collaborative settings where the objectives are static. On the other hand, with the Pursuit-Evasion scenario, we explored the utility of legible behaviours in collaborative-competitive settings where a team's objectives, in our case capture all the preys, were in competition with those of other teams. Both of these experiments explore different facets of agent collaboration. 

The experimental results show that legible behaviours have a positive impact in tasks that require teamwork, with the impact being clearer in scenarios like the LB-Foraging scenario, rather than in scenarios like the Pursuit-Evasion where teams have conflicting objectives. In both scenarios, the teams with a legible leader outperformed teams with an optimal leader in settings with smaller grid spaces, requiring, on average, less steps to complete the task and for the follower agent to correctly infer the next objective. In larger grids, we observed that in cooperation tasks with full objective alignment, like in LB-Foraging, legible leaders continue the trend of leading the teams to outperform the teams that have optimal leaders. In contrast, in tasks with competing objectives, the larger state spaces give opposing agents more space to coordinate, making the task harder to solve and the difference between legible and optimal leaders less significative.

Overall, we can conclude that using legible behaviours leads to agents being better at conveying their intentions. The results show that legible decision making is good for applications where teamwork is required. Particularly, legible decision making is a good option in team settings where identifying other teammates objectives is paramount; such as in ad-hoc teamwork scenarios, where intelligent agents join teams mid task and need to infer what task the team is solving. Thus, by having legible agents on the team, new team members will understand the team's task quicker than if the team was composed solely by optimal agents.


However, in tasks where different teams of agents have competing or opposing objectives, it is important to have a balance between optimal and legible behaviours in order to achieve peak performance. In such contexts, a legible agent’s emphasis on transparency might not always be the best course of action, as it may give opposing teams time to organize and negate the advantages a legible leader provides of being more transparent to other teammates. Thus, we propose that future research might focus on researching mixed strategies of legible and optimal behaviours, taking advantage of the benefits the two behaviour strategies provide. An interesting approach for mixed strategies is allowing the legible agents to switch between optimal and legible behaviours, as needs arise. For example, in the Pursuit-Evasion scenario, a legible agent might use legible behaviours while the preys are more spaced, but, when clusters of preys begin to form, the agent switches to optimal behaviours.

The results of this study focus on the impact of legibility in teams of autonomous agents with a strict leader-follower hierarchy. However, in real-world scenarios these leader-follower dynamics are not always so strict or they might even be non-existent. So, we consider that it might be interesting, for future work, to explore collaborative tasks with a more horizontal hierarchy, where there is not a single leader that defines the pace of the task, but every agent can contribute to the task's pace and to how different objectives are tackled.



\section*{Acknowledgements}
This work was supported by the Portuguese Recovery and Resilience Plan via project C645008882-00000055 (Center for Responsible AI). Additional support was provided by FCT/MECI through national funds and, when applicable, co-funded EU funds under UID/50008: Instituto de Telecomunicações, and under project UIDB/50021/2020.



\bibliographystyle{unsrtnat}
\bibliography{references}


\end{document}